\documentclass{bmvc2k}

\title{Deep Multimodal Fusion Detection through Spatial Mask and Channel Competition}

\addauthor{Guandi Wang$^1$}{guandi@kth.se}{1}
\addauthor{Ming Li$^2$}{minglii@umd.edu}{2}
\addauthor{Yunsen Xing$^1$}{yunsen@kth.se}{1}
\addauthor{Junle Liu$^1$$\dagger$}{junle@kth.se}{1}

\addinstitution{
 1, KTH Royal Institute of Technology
 Stockholm, Sweden\\
2, University of Maryland, College Park, USA
}

\runninghead{Wang {\emph{et al}\bmvaOneDot}}{MM Detection via spatial mask and channel competition}

\usepackage{amssymb}
\usepackage{float}
\usepackage{booktabs}
\usepackage{multirow}
\usepackage{wrapfig}
\usepackage[table]{xcolor}
\usepackage{needspace}
\definecolor{clrFull}{HTML}{FAEEDA}
\definecolor{clrLCC}{HTML}{F1EFE8}

\begin{document}
\maketitle
\let\thefootnote\relax\footnotetext{$\dagger$ Corresponding Author}

\begin{abstract}
Deep multimodal fusion for object detection has demonstrated good performance through mining modal characteristics. However, existing feature-level fusion methods mainly weigh between two modalities and unify them in a unified representation space. This can lead to overfitting or over-specialization of the statistical properties of a single modality within a dual-backbone architecture. This paper proposes an Attention-Driven Complementarity Resampling framework for robust improvement of cross-modality object detection. Based on a shared channel spatial attention mechanism, we first introduce the semantic mask exchange to actively mix the boundaries of the modalities during the training phase, forcing the backbone network to learn generalized features without relying on fixed modal labels. Then we propose a learnable channel competition to sample and aggregate features in a channel-wise and learnable way. Our experiments on multiple datasets demonstrate that the proposed method is effective and yields competitive results among existing state-of-the-art approaches.
\end{abstract}

\section{Introduction}
\label{sec:intro}
Multimodal object detection, specifically leveraging visible and infrared data, has emerged as a critical component for robust perception systems in surveillance~\cite{10.1145/3545572} and autonomous 
driving~\cite{lin20223d}. The core advantage of integrating visible and thermal information lies in 
the complementary nature of different modalities: RGB sensors provide rich texture and color details in favorable lighting conditions, while infrared sensors capture thermal signatures impervious to illumination 
changes or camouflage. Driven by the demand for robust perception, existing fusion methods have evolved along 
two primary trajectories: shallow heuristic aggregation, which relies on designed weights or gating mechanisms such as addition and concatenation, and parametric dense aggregation, which employs cross-attention~\cite{shen2024icafusion}~\cite{qingyun2021cross} or state space models~\cite{dong2025fusion} to map heterogeneous features into a joint representational space.

Despite their differences in complexity, existing methods share a common underlying assumption that the two 
modalities can and should be harmonized into a unified 
representation by exploiting their respective modal 
characteristics~\cite{li2018densefuse}. However, visible and infrared images are 
fundamentally heterogeneous signals, and they arise from 
entirely different physical sensing mechanisms and 
encode scene information in intrinsically different ways~\cite{ma2019infrared}. 
Directly harmonizing such heterogeneous signals through feature mixing, whether via shallow weighting or deep attentional entanglement, is essentially a passive compromise that treats modality characteristics as the primary basis for fusion decisions.

This observation motivates a fundamentally different perspective on multimodal fusion mechanism. Rather than harmonizing modality-specific characteristics, we explore the mechanism that the networks were trained to make fusion decisions without relying on modality identity.  Inspired by the regularization success of spatial augmentation strategies such as CutMix~\cite{yun2019cutmix} in unimodal settings, we propose to internalize a semantically guided feature exchange mechanism within the backbone network, actively circulating modality-specific spatial features between branches during training. This partial mixing of modality sources can reduce the network's dependence on the inherent characteristics of a particular modality. Consequently, the downstream fusion module can perform pure channel-level competition without needing to reason about modality characteristics, selecting the more informative channel regardless of its modal origin.

Based on this observation, we propose Attention-Driven Complementary Resampling 
(ADCR), a training-inference decoupled fusion framework consisting of two complementary modules. The Semantic Mask Enhancement 
(SME) module operates exclusively during training, 
performing semantics-guided spatial feature exchange 
between modality branches to prevent over-specialization 
to single-modality statistics. The Learnable Channel 
Competition (LCC) module then performs channel-level 
arbitration at inference via a differentiable competition 
mechanism, selecting the more reliable channel from each 
modality without additional supervision signals. Together, SME and LCC form a 
principled decoupling of the where-to-complement and 
how-to-aggregate sub-problems in multimodal fusion.
The main contributions of this work are as follows:

\begin{itemize}
    \item We propose a training-time semantically-guided spatial exchange mechanism that prevents backbone over-specialization by actively circulating cross-modal features during training, with zero inference overhead.
    
    \item We propose an inference-time channel competition module that performs modality-agnostic channel arbitration  via a differentiable Entmax-based mechanism, requiring no additional loss functions.
    
    \item We conduct comprehensive experiments on five 
    public benchmarks LLVIP, M$^3$FD, FLIR, VEDAI, 
    and DroneVehicle, demonstrating consistent 
    improvements over state-of-the-art (SOTA) methods, with particularly substantial gains on datasets characterized by severe modality asymmetry and 
    spatial misalignment.
\end{itemize}

\section{Related Work}
\subsection{Multi spectral Information Fusion}
Multi-spectral information fusion has shown considerable advantages in practical engineering applications, like autonomous driving, robotics and drones~\cite{ma2019infrared,hwang2015multispectral}. Recent advancements in multimodal object detection have increasingly adopted Transformer-based architectures to model long-range dependencies~\cite{carion2020end,bai2022transfusion,ma2022swinfusion}. Methods~\cite{qingyun2021cross,shen2024icafusion} leverage multi-head self-attention mechanisms to interactively fuse features across different scales, achieving significant performance gains over traditional convolutional neural networks(CNN)-based fusion. To further refine the fusion process, some works have focused on introducing specific inductive biases into the specific mechanism in their fusion paradigms. For instance, CrossFuse~\cite{li2024crossfuse} addresses the potential redundancy in standard attention by incorporating a reverse softmax constraint, which explicitly guides the network to focus on complementary information between modalities. 
Similarly, IF-USOD~\cite{yuan2025if} introduces a cross-scale interactive embedding bias, utilizing a hybrid CNN-transformer structure to facilitate effective information flow between high-level semantics and low-level details. 
Though existing methods have successfully enhanced feature representation through dense interactions, they generally model scale variations and modality characteristics in a coupled manner. 
This leads to a passive compromise that ignores the fundamental physical heterogeneity between thermal radiation and reflected light~\cite{zhao2023cddfuse}.

However, a notable challenge for these coupled mixing strategies emerges in real-world scenarios characterized by significant semantic asymmetry, such as dense smoke or extreme low-light conditions~\cite{liu2022target, jia2021llvip}. When the information distribution between modalities is highly imbalanced, direct feature mixing often results in an undesirable feature compromise. Rather than explicitly identifying and compensating for spatial representation deficiencies, dense attention mechanisms may inadvertently propagate unreliable cues from the less informative modality into the robust counterpart. This interaction tends to attenuate the sharp, high-frequency edges that are crucial for precise bounding box regression~\cite{lin20223d}. Therefore, exploring disentangled paradigms that explicitly diagnose and correct spatial imbalances before performing lightweight, parameter-efficient aggregation remains a highly valuable research direction.

\subsection{Cross-Modal Feature Exchange and Regularization}

To achieve explicit spatial correction without the massive computational overhead of dense attention, alternative approaches have explored non-parametric feature exchange. Inspired by the regularization success of multimodal spatial augmentations like CutMix~\cite{yun2019cutmix}, recent multispectral methods have introduced cross-modal swapping mechanisms.
CDC-YOLO Fusion~\cite{wang2024cdc} and LRAF-Net~\cite{fu2023lraf} propose image-level data augmentation strategies that randomly exchange or mask local patches between visible and infrared inputs to force multimodal learning. Concurrently, at the feature level, methods like Channel Exchange Networks~\cite{wang2020deep} and the State Space Channel Swapping module in FusionMamba~\cite{dong2025fusion} perform fixed exchanges along the channel dimension.

Though previous works mainly focus on predefined exchange patterns and symmetric feature interaction, refining the exchange granularity worth exploring. Extending the concept of image-level swapping into the deep feature extraction phase offers a promising pathway to dynamically capture semantic imbalances. Furthermore, inspired by the interaction promoted by channel-level swapping, shifting towards a spatially targeted exchange can effectively prevent the propagation of unreliable cues under asymmetric degradation. Motivated by these insights, we aim to design a spatially-aware, directional exchange strategy that fosters deep cross-modal co-adaptation while meticulously preserving representation integrity.

\section{Method}
\label{sec:method}
In multimodal dense prediction, the objective of fusion is to integrate complementary information while suppressing noise from a specific modality. Given a pair of input features $F_{vis}, F_{ir} \in \mathbb{R}^{C \times H \times W}$, a conventional fusion function $\Phi(\cdot)$ aims to produce a unified representation:

\begin{equation}
    F_{fused} = \Phi(F_{vis}, F_{ir}).
\end{equation}

Under the attention-weighted feature transformation
$F' = F \odot \mathcal{W}$, the attention weight $\mathcal{W} \in (0,1)$ directly scales the gradient flow to the input feature, making it a natural proxy for information density rather than a mere mixing coefficient: regions where features are task-irrelevant or corrupted are naturally suppressed through optimization. However, we observe a fundamental optimization conflict in static fusion paradigms. 
In scenarios where both modalities are reliable, the fusion paradigm $\Phi$ should achieve comprehensive information retention to capture subtle complementary details across modalities. Conversely, under asymmetric modal performance, where one modality is corrupted by environmental noise, ideally $\Phi$ should function as a principled isolator to binary-truncate the compromised pathways and prevent noise leakage. Forcing a single mapping function to simultaneously satisfy these two divergent objectives often leads to gradient ambiguity and sub-optimal convergence.

\begin{figure*}[htb!]
\begin{center}
\includegraphics[width=1\linewidth]{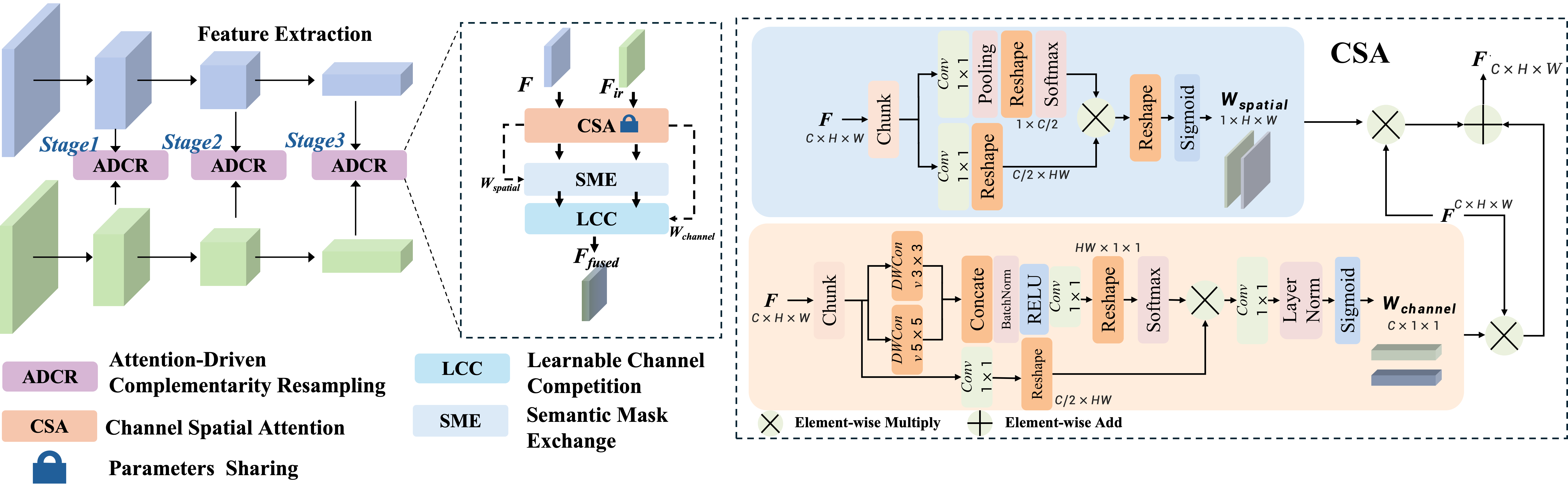}  
\end{center}
   \caption{\textbf{Pipeline of our proposed method.} The dashed arrows indicate control flow and the solid arrows indicate the direction of network feature transmission. We select a dual-backbone feature extraction network as the backbone, and ADCR modules are embedded in three different stages of the network. The two branches do not share their parameters; the channel spatial attention is shared for two modalities. The CSA module employs a parallel two-branch architecture: the spatial branch generates a spatial mask $\textbf{W}_{spatial}$ through $1\times1$ convolutions and tensor multiplication; the channel branch introduces parallel $3\times3$ and $5\times5$ depthwise separable convolutions to extract multi-scale context, generating a channel mask $\textbf{W}_{channel}$ while smoothing potential spatial misalignments. Finally, the original features are element-wise multiplied with both masks and then summed to output a doubly refined unified representation $F'$.
}
\label{fig:intro}
\end{figure*}

As shown in \textbf{Fig.}\ref{fig:intro}, we design our proposed Attention-Driven Complementarity Resampling module in three stages of a dual-backbone network. Firstly, we propose a modality-sharing Channel-Spatial Attention (CSA) to extract spatial and channel weights as the theoretical confidence metric $\mathcal{W}$ into a computable and cross-modal comparable scale. Given the visible feature  
$F_{vis}\in \mathbb{R}^{C \times H \times W}$ and infrared feature $F_{ir} \in \mathbb{R}^{C \times H \times W}$, both modalities are projected into a unified geometric space through the Spatial Attention (SA) module with identical convolutional weights and normalization parameters. Through channel-only and spatial-only branches, we extract the channel confidence mask $\mathcal{W}_{ch} \in \mathbb{R}^{C \times 1 \times 1}$ and the spatial confidence mask $\mathcal{W}_{sp} \in \mathbb{R}^{1 \times H \times W}$. 

Heterogeneous modalities such as visible and infrared images exhibit fundamentally different photometric distributions. 
By enforcing identical convolutional weights and normalization statistics across both modalities, the CSA projects $F_{vis}$ and $F_{ir}$ into a unified geometric space where the sole source of variation in $\mathcal{W}$ is the information content of the input rather than parametric bias. Under this constraint, the discrepancy between $\mathcal{W}_{vis} $ and $ \mathcal{W}_{ir}$ reflects purely content-driven divergence, enabling the subsequent methods to operate on a dimensionless, cross-modal comparable scale. 

\subsection{Semantic Mask Exchange via Spatial-Semantic Discrepancy}
\label{sec:method_2}

$\mathcal{W}_{ch}$ and $\mathcal{W}_{sp}$ serve to quantify intra-modal feature utility. By comparing these confidence maps across modalities, we can identify regions where information density is spatially imbalanced. In regions where one modality exhibits overwhelming superiority, characterized by a large gap between $\mathcal{W}_{vis,sp}$ and $\mathcal{W}_{ir,sp}$, a structured cross-modal spatial exchange can enrich feature representation of the weaker modality. We propose SME to utilize this insight during training as a complementarity-driven spatial augmentation.

\begin{figure*}[htb!]
\begin{center}
\includegraphics[width=0.9\textwidth]{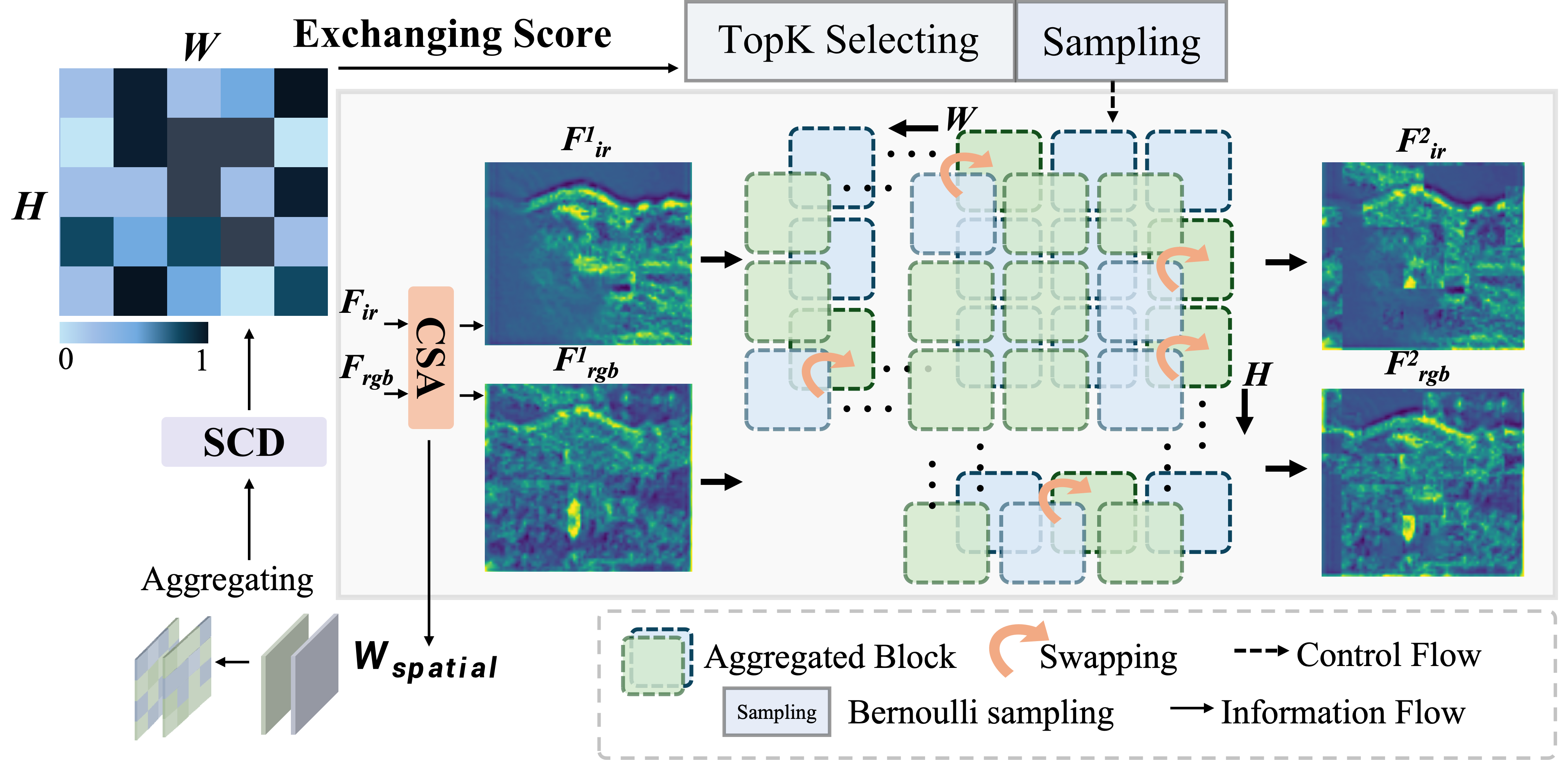}  
\end{center}
   \caption{\textbf{Semantic Mask Exchange Mechanism.}
The exchange score ($N \times N$)is calculated by Semantic Cosine Distance (SCD), and is subsequently sparsified through a top-$K$ selection.
The exchange is operated during the training stage, and we set a Bernoulli sampling to regularize the distribution of exchanged features. It utilizes continuous exchange scores $s_{i,j}$ as activation probabilities, ensuring that the fusion boundary remains fluid while prioritizing the exchange of highly complementary regions. 
}
\label{fig:SME}
\end{figure*}

As illustrated in \textbf{Fig.}\ref{fig:SME}, we first discretize the spatial confidence maps into a block-level representation.
Given $\mathcal{W}$ in blocks and input weight $\mathcal{W}_{sp} \in \mathbb{R}^{B \times 1 \times H \times W}$, we aggregate global features into a block-level representation in size $B \times N \times N$. Let the height of each block be $h = \lfloor H/N \rfloor$, and the width be $w = \lfloor W/N \rfloor$. For the $b$-th sample in the batch, the weight $V_{b, i, j}$ of its aggregated $(i, j)$-th block can be expressed as:

\begin{equation}
   V = \frac{1}{C \cdot h \cdot w} \sum_{c=1}^{C} \sum_{x=i \cdot h}^{(i+1)h-1} \sum_{y=j \cdot w}^{(j+1)w-1} W_{b, c, x, y}
\end{equation}

For each block $(i,j)$, we define the exchange score as Semantic Cosine Distance(SCD), a
product of two complementary terms:
\begin{equation}
    s_{i,j} =
    \underbrace{
        \frac{\bigl|V_{ir}^{(i,j)} - V_{vis}^{(i,j)}\bigr|}
             {2(\max_{m,n}\bigl|V_{ir}^{(m,n)}-V_{vis}^{(m,n)}\bigr|+\epsilon)}
    }_{\text{Spatial Attention Gap}}
    \cdot
    \underbrace{
        \left(1 - \frac{\mathbf{f}_{ir}^{(i,j)} \cdot \mathbf{f}_{vis}^{(i,j)}}
                {\|\mathbf{f}_{ir}^{(i,j)}\|_2\,\|\mathbf{f}_{vis}^{(i,j)}\|_2}
        \right)
    }_{\text{Semantic Cosine Distance}}
\end{equation}
where:
\begin{itemize}
    \item \textbf{Semantic Cosine Distance} adaptively adjusts feature representation during the fusion process by measuring the semantic consistency of features between modalities. When visible light and infrared features are highly homogeneous, the distance approaches 0, thus fundamentally suppressing meaningless feature oscillations; conversely, when there is strong inconsistency between modalities, their feature descriptors will exhibit high orthogonality in the semantic space, and the distance will approach 1.
    
    \item \textbf{Spatial Attention Gap} is smoothed and continuously modulated. $\mathbf{f}^{(i,j)}$ denotes the block-level feature vector obtained by spatially averaging the weighted features within block $(i,j)$. It can simultaneously represent the degree of difference in activation features between modes and the difference in cross-modal features of spatial blocks $(i,j)$ in the network. $(m,n)$ represents traversing all local grids of the current feature map.
\end{itemize}

Given the exchange scores $\{s_{i,j}\}$, we select the top-$K$ blocks as candidates and apply stochastic selection via Bernoulli sampling~\cite{srivastava2014dropout}:

\begin{equation}
    m_{i,j} \sim
\mathrm{Bernoulli}\!\bigl(s_{i,j}\cdot\mathbb{I}\!\left((i,j)\in\mathcal{I}_K\right)\bigr),
\end{equation}
where $\mathcal{I}_K$ denotes the index set of the top-$K$ blocks
ranked by $s_{i,j}$.

For selected blocks, the corresponding spatial regions of
$F_{vis}$ and $F_{ir}$ are symmetrically swapped, yielding the
spatially mixed features $\tilde{F}_{vis}$ and $\tilde{F}_{ir}$.
Semantic Mask Exchange (SME) is applied exclusively during training with a probability of $p$, enforcing each modality branch to incorporate complementary spatial context from the other modality and preventing over-specialization. In the inference phase, SME is deactivated with no additional computational overhead.

\subsection{Learnable Channel Competition}
From an optimization perspective, the parallel maintenance of dual-modality channels creates bifurcated gradient pathways. For any channel $c$ where both modalities encode redundant information, the optimizer receives coupled gradient signals, leading to feature co-adaptation and inefficient representation; when one modality is corrupted, its gradient path continues to propagate noise into the shared backbone. A channel-level competition mechanism collapses these bifurcated paths into a single, trustworthy gradient conduit per channel.

\begin{figure*}[htb!]
\begin{center}
\includegraphics[width=0.9\linewidth]{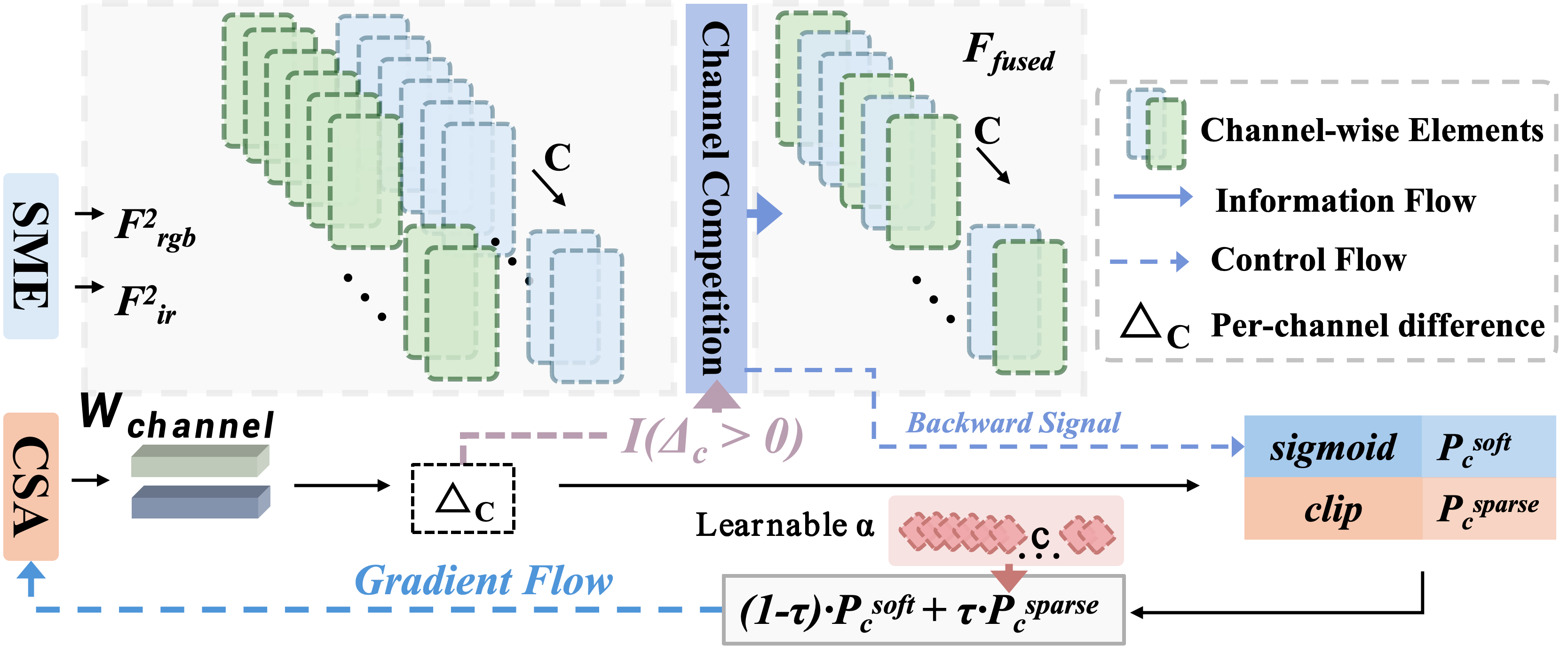}  
\end{center}
   \caption{\textbf{Learnable Channel Competition Mechanism.} Previous spatially mixed features undergo per-channel arbitration based on the inter-modal confidence discrepancy $\Delta_c$.
The channel competition branch is controlled by an indicator function  $I(\Delta_c > 0)$ that executes a discrete decision. To overcome the non-differentiability of the indicator function, an Adaptive Boundary Proxy is constructed using a convex interpolation of a soft sigmoid ($P_c^{soft}$) and a sparse clip ($P_c^{sparse}$) component, regulated by a learnable parameter $\alpha_c$. The adaptive boundary proxy (ABP) allows the error gradients to intelligently bypass the discrete indicator and flow exclusively through the continuous proxy, ensuring end-to-end optimizability for the upstream modules.
}
\label{fig:LCC}
\end{figure*}

Let $\tilde{F}_{vis}, \tilde{F}_{ir} \in \mathbb{R}^{B \times C \times H \times W}$
be the spatially mixed features from SME.
For each channel $c$, we define a binary selection variable
$z_c \in \{0,1\}$, where $z_c = 1$ retains the visible modality and
$z_c = 0$ retains the infrared modality.
The fused output is expressed as:

\begin{equation}
    F_{:, c, :, :}^{fuse} = z_c \cdot \tilde{F}_{vis, :, c, :, :} + (1 - z_c) \cdot \tilde{F}_{ir, :, c, :, :}.
\end{equation}

Based on the unified projection of CSA, these weights are inherently comparable across modalities, allowing us to define the channel-level discrepancy as:

\begin{equation}
    \Delta_c = \mathcal{W}_{vis,ch}^{(c)} - \mathcal{W}_{ir,ch}^{(c)}.
\end{equation}

Following the ordinal interpretation, $\Delta_c > 0$ 
indicates higher information density in the visible modality, while $\Delta_c < 0$ favors the infrared stream. Ideally, $z_c = \mathbb{I}(\Delta_c > 0)$, but this indicator function yields zero gradients almost everywhere, severing the backpropagation path to the CSA parameters and $\alpha$. To restore differentiability while preserving hard forward-pass decisions, we construct an adaptive boundary proxy(ABP) $\pi_c^{vis} \in [0,1]$ to approximate the selection probability of the visible modality. It is formulated as a convex interpolation between a \emph{soft} component and a \emph{sparse} component. The soft component uses a scaled sigmoid to maintain dense gradient flow:

\begin{equation}
    p_c^{soft} = \sigma(s \cdot \Delta_c),
\end{equation}
where $s$ is a predefined logit scale. The sparse component uses a piece-wise linear clip that maps directly
to $\{0,1\}$ outside the sensitive region $[-1/s,\,1/s]$:

\begin{equation}
    p_c^{sparse}
    = \mathrm{clip}\!\left(0.5 + \tfrac{s \cdot \Delta_c}{2},\; 0,\; 1\right).
\end{equation}

A per-channel learnable parameter $\alpha_c \in [1.01, 2.0]$ controls the interpolation stiffness via $\tau_c = \alpha_c - 1$:

\begin{equation}
    \pi_c^{vis}
    = (1-\tau_c)\cdot p_c^{soft} + \tau_c \cdot p_c^{sparse}.
\end{equation}

Inspired by $\alpha$-Entmax~\cite{peters2019sparse}, the ABP will function as below:
$\tau_c \rightarrow 0$, where the proxy reduces to sigmoid (dense gradients); 
as $\tau_c \rightarrow 1$,
it approaches sparsemax (sparse, boundary-polarized gradients). 
This adaptation ensures that the selection remains differentiable while possessing the structural capacity for absolute truncation. The indicator function is piecewise constant and yields zero gradients almost everywhere, and it effectively severs the backpropagation path, preventing the joint optimization of the CSA parameters and the entropy modulator $\alpha_c$. To bridge this gap between discrete semantic arbitration and continuous gradient descent, we employ the Straight-Through Estimator~\cite{bengio2013estimating}. We formulate the final routing mask $\mathcal{M}_c^{vis}$ by decoupling its forward-pass value from its backward-pass derivative as:

\begin{equation}
    z_c
    = \underbrace{\mathbb{I}(\Delta_c > 0)}_{\text{forward}}
    - \,\mathrm{sg}(\pi_c^{vis})
    + \pi_c^{vis},
\end{equation}

where $\text{sg}(\cdot)$ denotes the stop-gradient operator. During the forward pass, $z_c$ is numerically identical to the discrete decision $\mathbb{I}(\Delta_c>0)$. During backpropagation, gradients bypass the non-differentiable indicator and flow through $\pi_c^{vis}$ as:

\begin{equation}
    \frac{\partial \mathcal{L}}{\partial \Delta_c}
    = \frac{\partial \mathcal{L}}{\partial z_c}
      \cdot \frac{\partial \pi_c^{vis}}{\partial \Delta_c}.
\end{equation}

In inference, the soft proxy is discarded; only the hard decision $z_c = \mathbb{I}(\Delta_c > 0)$ is retained.

\section{Experiment}

\paragraph{Dataset.} To comprehensively evaluate the performance of our proposed methods, we conduct experiments on five public benchmarks: M$^3$FD~\cite{liu2022target}, LLVIP~\cite{jia2021llvip}, 
FLIR~\cite{farooq2021object},
DroneVehicle~\cite{sun2022drone},
and VEDAI~\cite{razakarivony2016vehicle}.

\paragraph{Implementation.} Our method uses untrained CSPdarknet~\cite{bochkovskiy2020yolov4} as the backbone for feature extraction. The training was conducted on a \textit{NVIDIA A40} GPU with \textit{the SGD} optimizer; the environment setting is \textit{PyTorch-2.5.0-NGC-24.09}. 
The spatial partition number $N$ and top-$K$ selection ratio in SME are fixed to 10 and 50\% respectively throughout all experiments. It ensures that each block contains sufficient spatial context information covers a wide range of candidate swap regions, while avoiding redundant swaps in highly similar regions. 
The batch size is set as 16, and we ran 200 epochs to obtain the final result. The training hyperparameters are set as follows: learning rate is set as 0.01 for both initial learning rate and final learning rate, and momentum is set as 0.937.

\subsection{Comparison with other methods}

We evaluate our method on five benchmark datasets against both single-modality baselines and multi-modal fusion methods. 

On the dataset of LLVIP and M$^3$FD, shown in \textbf{Tab.}~\ref{tab:sota_main}, our ADCR model achieves competitive performance, indicating ADCR outperforms other models in pedestrian and multi-class detection under relatively controlled capture conditions. On LLVIP, the CSPDarknet53-v8 backbone variant achieves 65.1\% mAP$_{95}$, with a 0.2\% gain over the second-best method. On M$^3$FD, ADCR reaches 88.2\% mAP$_{50}$ and 61.0\% mAP$_{95}$, consistent with the trend that decoupled channel reasoning offers steady improvement even when modality quality is relatively balanced.
In particular, the advantage becomes more pronounced when modality asymmetry is inherent to the capture conditions. On FLIR, where unstable visible-light quality causes infrared to dominate scene semantics, our ADCR model achieves 79.1\% mAP$_{50}$ and 43.0\% mAP$_{95}$, exceeding the next best result by 0.9\% and 3.1 points, respectively. This denotes that LCC's inference-time channel arbitration is particularly effective when one modality consistently contributes more reliable cues, allowing the model to selectively amplify rather than uniformly fuse cross-modal information.

\begin{table*}[htb]
\centering
\caption{Comparison with state-of-the-art methods across four benchmarks. \textbf{Bold} and \underline{underline} indicate the best and second-best performances.
}
\label{tab:sota_main}
\setlength{\tabcolsep}{4pt}
\renewcommand{\arraystretch}{1.0}
\footnotesize
\resizebox{0.9\textwidth}{!}{%
\begin{tabular}{l l cc cc cc cc}
\toprule
\multirow{2}{*}{Method} & \multirow{2}{*}{Backbone} &
  \multicolumn{2}{c}{LLVIP} &
  \multicolumn{2}{c}{M$^3$FD} &
  \multicolumn{2}{c}{FLIR} &
  \multicolumn{2}{c}{VEDAI} \\
\cmidrule(lr){3-4}\cmidrule(lr){5-6}\cmidrule(lr){7-8}\cmidrule(lr){9-10}
& & mAP$_{50}$ & mAP$_{95}$ & mAP$_{50}$ &  mAP$_{95}$ & mAP$_{50}$ &  mAP$_{95}$
  & mAP$_{50}$ &  mAP$_{95}$ \\
\midrule

\multicolumn{10}{l}{\textit{Single-modality baselines}} \\[1pt]
YOLOv5~\cite{jocher2021yolov5} (RGB) & CSPDarknet53-v5
  & 90.8 & 50.0 & 83.5 & 55.2 & 67.8 & 31.8 & 74.3 & 64.2 \\
YOLOv5~\cite{jocher2021yolov5} (IR)  & CSPDarknet53-v5
  & 94.6 & 61.9 & 84.2 & 56.1 & 73.9 & 39.5 & 74.0 & 64.2 \\
YOLOv8~\cite{ultralytics_yolo} (RGB) & CSPDarknet53-v8
  & 91.9 & 54.0 & 80.9 & 52.5 & 66.3 & 28.2 & 52.1 & 31.0 \\
YOLOv8~\cite{ultralytics_yolo} (IR)  & CSPDarknet53-v8
  & 95.2 & 62.1 & 79.5 & 53.1 & 72.9 & 38.3 & 56.9 & 35.3 \\
\midrule

\multicolumn{10}{l}{\textit{Multi-modal fusion methods}} \\[1pt]
GAFF~\cite{zhang2021guided}          & ResNet18
  & 94.0 & 55.8 & --   & --   & 72.7 & 37.3 & 72.9 & 37.5 \\
CFT~\cite{qingyun2021cross}          & CSPDarknet53-v5
  & 96.8 & 62.9 & 83.2 & 52.2 & 74.5 & 36.5 & 74.6 & 44.1 \\
CFR~\cite{zhang2021guided}              & CSPDarknet53-v5
  & --   & --   & 82.7 & 50.3 & 72.4 & --   & --   & --   \\
ICAFusion~\cite{shen2024icafusion}   & CSPDarknet53-v5
  & 97.0 & 62.4 & 86.6 & 56.9 & 77.3 & 38.1 & 75.7 & 46.2 \\
MCOR~\cite{jang2025multispectral}    & CSPDarknet53-v5
  & \underline{97.6} & \underline{64.9}
  & 87.2 & 57.3
  & \underline{78.2} & 39.9
  & 76.2 & 46.3 \\
TarDAL~\cite{liu2022target} & CSPDarknet53-v5
& -- & -- & 78.0 & 46.9
  & -- & -- & -- & -- \\
CMAFF~\cite{qingyun2022cross}        & CSPDarknet53-v5
  & -- & -- & -- & --
  & -- & -- & 68.0 & 42.6 \\
Fusion-Mamba~\cite{dong2025fusion}         & CSPDarknet53-v5
  & 96.8 & 62.8 & 85.0 & 57.5
  & -- & -- & -- & -- \\
Fusion-Mamba~\cite{dong2025fusion}         & CSPDarknet53-v8
  & 97.0 & 64.3
  & \underline{88.0} & \textbf{61.0}
  & -- & -- & -- & -- \\
\midrule
ADCR (Ours) & CSPDarknet53-v8
  & 97.5 & \textbf{65.1}
  & \textbf{88.2} & \textbf{61.0}
  & \textbf{79.1} & \textbf{43.0}
  & \underline{82.7} & 49.5 \\
ADCR (Ours) & CSPDarknet53-v5
  & \textbf{97.7} & 64.2
  & 85.9 & \underline{58.6}
  & \underline{78.2} & \underline{41.8}
  & 81.9  & \underline{50.2} \\
\bottomrule
\end{tabular}
}
\end{table*}

\begin{table*}[h!]
\centering
\caption{ Comparison with state-of-the-art methods on the DroneVehicle benchmark.
  \textbf{Bold} and \underline{underline} indicate the best and second-best results.}
\label{tab:dronevehicle}
\setlength{\tabcolsep}{5pt}
\renewcommand{\arraystretch}{1.0}
\footnotesize
\resizebox{0.5\textwidth}{!}{
\begin{tabular}{l l c c}
\toprule
Method & Backbone & mAP$_{50}$ & mAP \\
\midrule
\multicolumn{4}{l}{\textit{Single-modality baselines}} \\[1pt]
YOLOv5~\cite{jocher2021yolov5} (RGB)        & CSPDarknet53-v5 & 65.2 & 37.7 \\
YOLOv5~\cite{jocher2021yolov5} (IR)        & CSPDarknet53-v5 & 76.9 & 53.3 \\
YOLOv8 (RGB)~\cite{ultralytics_yolo}       & CSPDarknet53-v8 & 71.7 & 43.3 \\
YOLOv8 (IR)~\cite{ultralytics_yolo}        & CSPDarknet53-v8 & 80.4 & 57.6 \\
\midrule
\multicolumn{4}{l}{\textit{Multi-modal fusion methods}} \\[1pt]
AR-CNN~\cite{zhang2019weakly}                & 
VGG-16          & 71.6 & 40.5 \\
UA-CMDet~\cite{sun2022drone}              & ResNet50        & 64.0 & 48.2 \\
TSFADet~\cite{yuan2022translation}                 & ResNet50        & 73.1 & 44.1 \\
SLBAF-Net~\cite{cheng2023slbaf}             & CSPDarknet53-v5 & 77.0 & 49.5 \\
CSOM-ODAF~\cite{chen2024weakly}               & CSPDarknet53-v5 & 77.5 & 52.5 \\
CMAFF~\cite{qingyun2022cross}        & CSPDarknet53-v5 & 82.0 & 57.6 \\
CMA~\cite{jiang2024m2fnet}                      & CSPDarknet53-v8 & 76.8 & 52.4 \\
Fusion-Mamba~\cite{dong2025fusion}     & CSPDarknet53-v8 & \underline{79.2} & \underline{56.0} \\
\midrule
\textbf{ADCR (Ours)}                        & CSPDarknet53-v5      & \textbf{84.4} & \textbf{64.5} \\
\textbf{ADCR (Ours)}                        & CSPDarknet53-v8      & \textbf{84.1} & \textbf{64.3} \\
\bottomrule
\end{tabular}
}
\end{table*}

Our method also have competitive performance on DroneVehicle, shown as \textbf{Tab.}~\ref{tab:dronevehicle}, where aerial viewpoints introduce compounded challenges: large-scale spatial misalignment and severe modality-asymmetric appearance variation co-occur across the same scene. ADCR achieves leading results with two backbone. We attribute this to the complementary effect of SME's training-time spatial resampling, which encourages robustness to geometric inconsistency, and LCC's inference-time arbitration, as further validated in the ablation study (\textbf{Tab.}~\ref{tab:ablation}). On VEDAI, ADCR surpasses all evaluated multi-modal fusion methods with 82.7\% mAP$_{50}$ and 49.5\% mAP$_{95}$, further confirming the generalizability of the decoupled design across aerial detection scenarios.

\subsection{Ablation Experiments}

To validate the effectiveness of the proposed modules in the model, we conduct an evaluation on the FLIR dataset. As shown in \textbf{Tab.}~\ref{tab:ablation}, we evaluate each component's effectiveness. To justify the design of LCC, we compare four alternative 
fusion strategies of increasing complexity. \textbf{Add} 
performs direct element-wise summation without any selection 
mechanism. \textbf{ConcatConv} concatenates the two modality 
features and applies a $1\times1$ convolution for linear channel mixing. LCC(w/o ABP) is the basic architecture without the adaptive boundary proxy(ABP) mechanism, and  \textbf{CompeteFusion} is the method that directly conducts channel competition without gradient backpropagation and other components.

\begin{table*}[htb!]
\centering
\caption{\textbf{Ablation and inference efficiency results.}
The baseline uses element-wise addition. Added Params denotes the additional parameters relative to the baseline. FLOPs and FPS are measured under the same inference protocol on an NVIDIA A40. The colored section in table is detailed ablation evaluation for LCC. }
\label{tab:ablation}
\small
\setlength{\tabcolsep}{4pt}
\renewcommand{\arraystretch}{0.79}
\resizebox{\textwidth}{!}{
\begin{tabular}{lcc|ccc}
\toprule
Variant &
mAP$_{50}$ (\%) &
mAP$_{95}$ (\%) &
Added Params (M) &
GFLOPs &
FPS \\
\midrule

Baseline (Add)
& 77.3 & 41.9
& -- & -- & -- \\

CSA only
& 77.6 & 41.3
& 1.090 & 21.25 & 677.2 \\

w/o SME
& 78.2 & 42.6
& 1.091 & 21.27 & 665.6 \\

w/o LCC (Add)
& 78.1 & 41.4
& 1.090 & 21.25 & 703.3 \\

ADCR (full)
& 79.1 & 43.0
& 1.091 & 21.27 & 670.8 \\

\midrule
\rowcolor{clrLCC}
w/o LCC (ConcatConv)
& 77.5 & 42.1
& 3.187 & -- & -- \\

\rowcolor{clrLCC}
LCC (CompeteFusion Only)
& 78.7 & 42.6
& -- & -- & -- \\

\rowcolor{clrLCC}
LCC (w/o ABP)
& 78.7 & 42.5
& -- & -- & -- \\

\bottomrule
\end{tabular}}
\end{table*}

Our evaluation basically focuses on two proposed modules: the SME and 
the LCC, 
The results are 
presented in Tab.~\ref{tab:ablation}, where w/o $m$ denotes 
the model without part $m$.

Effects of SME and LCC modules: we recorded the results of removing the SME and LCC modules. When SME is removed, the network's performance decreases by 0.9\% in mAP$_{50}$ and 0.4\% in mAP$_{95}$, respectively. 
This is due to the lack of spatial interaction between modalities, and 
SME encourages robustness to spatial misalignment during training.
When LCC is removed, the performance decreases by 1.0\% and 1.6\% in mAP$_{50}$ and mAP$_{95}$, respectively.LCC achieves physical truncation, effectively preventing noise propagation between modes. Simultaneously, LCC enables the entire network to be jointly optimized with adaptive polarization sparse parameters. Ultimately, the edge information required for high-precision positioning is preserved.
Each component contributes positively to the final performance. 

We also further evaluate internal designs in LCC.
Regarding another fusion strategy, we remove the channel competition and build ConcatConv instead, and the performance decreases to 77.5\% in mAP$_{50}$ and 42.1\% in mAP$_{95}$. Then we set CompeteFusion as the only fusion mechanism, the performance increases by 1.2\% and 0.5\% in mAP$_{50}$ and mAP$_{95}$ compared to ConcatConv. Based on CompeteFusion, we add the STE(straight-through estimator) architecture. Contrary to our expectation, while it reconnects the backward gradient flow, it yields no improvement in mAP$_{50}$ and mAP$_{95}$. Because the derivative of a sigmoid function never reaches strictly zero, it continuously leaks update signals to the suppressed modality branch during backpropagation and causes implicit cross-modal gradient interference. 
This fully justifies the necessity of our Adaptive Boundary Proxy (ABP) in the full ADCR model. By incorporating a piece-wise linear clip component, ABP zeroes out the gradients for decisively suppressed channels. This optimization sparsity perfectly aligns the backward gradient flow with the forward discrete routing, eliminating interference and driving the performance to peak bounds.
To comprehensively evaluate the efficiency of our design, we calculate the net additional parameter and computational overhead introduced by the fusion mechanisms, excluding the base parameters of the zero-reference backbone.Taking the deepest feature stage ($C=1024$) as a representative example, a conventional ConcatConv fusion operation incurs a substantial additional burden of 2.10M parameters and 26.84 GFLOPs relative to the baseline. In contrast, our full ADCR framework requires a net increment of  1.09M parameters and 21.27 GFLOPs, achieving a  20.8\% reduction in computational complexity. Furthermore, internal analysis shows that most of parameters and computational resources are dedicated to the necessary CSA spatial channel alignment. Built upon this aligned representation, our core cross-modal interaction mechanisms add virtually no overhead: SME introduces no learnable parameters and incurs no inference-time cost, as it is entirely bypassed during the forward pass, while LCC contributes merely 0.001M parameters with a negligible computational footprint of 19.66 MFLOPs, occupies 0.09\% of total. This confirms that ADCR achieves a good balance between foundational alignment and highly efficient cross-modal interaction.

\subsection{Sensitivity of parameters }

\begin{figure}[htbp]
    \centering
    \begin{minipage}{0.50\textwidth}
        \centering
        \includegraphics[width=\linewidth]{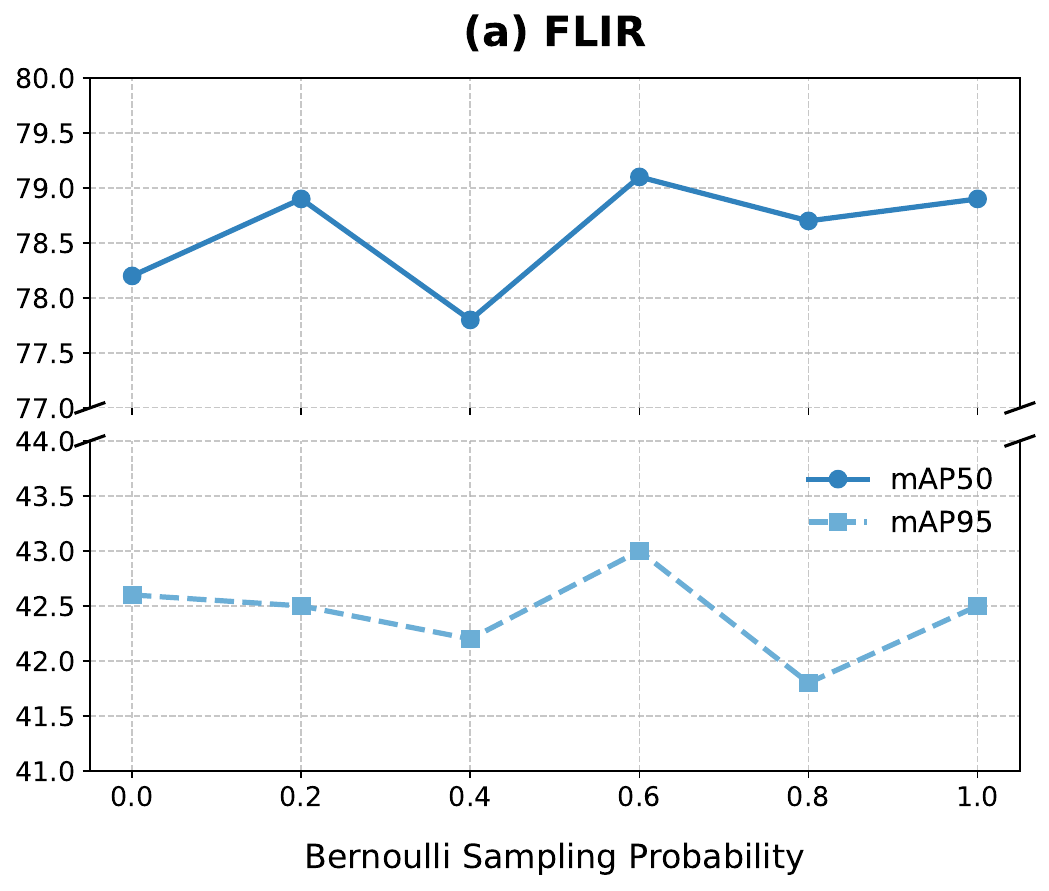}
    \end{minipage}\hfill
    \begin{minipage}{0.50\textwidth}
        \centering
        \includegraphics[width=\linewidth]{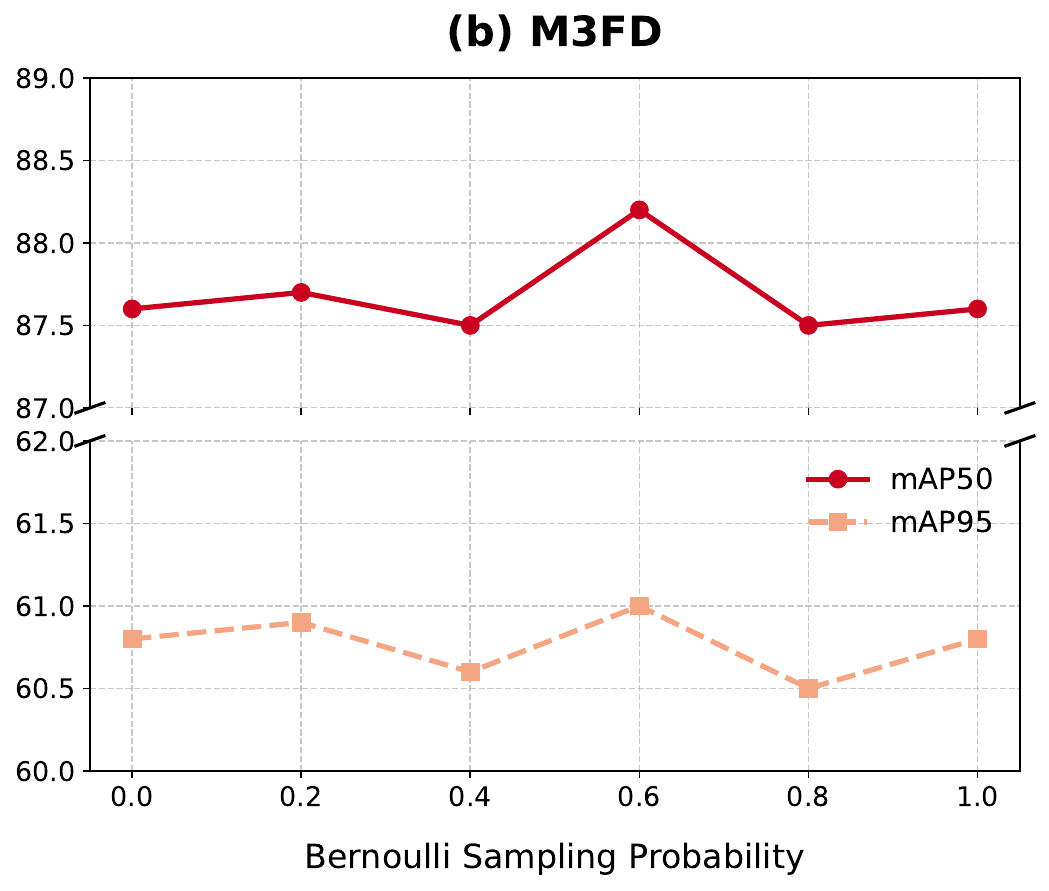}
    \end{minipage}
     \caption{\textbf{Sensitivity Analysis of the Bernoulli Sampling Probability.}
The consistent performance peak observed at the intermediate value ($p=0.6$) robustly validates the necessity of our proposed semantics-guided feature circulation during training.
     }
    \label{fig:sme_sensitivity}
\end{figure}

\paragraph{Bernoulli Sampling Probability.} As illustrated in Fig.~\ref{fig:sme_sensitivity}, we report the sensitivity analysis of the Bernoulli sampling probability $p$ across both the FLIR and M$^3$FD datasets to evaluate its impact on $mAP_{50}$ and $mAP_{95}$. The experimental curves reveal a consistent behavior trend between the two distinct benchmarks. Specifically, the detection performance does not change monotonically but peaks distinctively at $p=0.6$ on both datasets. The pattern presents that balanced cross-modal spatial exchange helps optimize cross-modal spatial regularizations without disrupting the learned backbone representations.

\par
\Needspace{12\baselineskip}
\begin{wraptable}{r}{0.36\columnwidth}
\centering
\small 
\setlength{\tabcolsep}{8pt}
\renewcommand{\arraystretch}{1.15}

\caption{\textbf{SME sensitivity to $N$ and $K$ on FLIR.}
Full ADCR is used in all experiments.}
\label{tab:nk_sensitivity}

\begin{tabular}{@{}ccc@{}}
\toprule
$N$ & $K$ & mAP$_{50}$/mAP$_{95}$ \\
\midrule
8  & 50\% & 78.6/42.6 \\
10 & 50\% & 79.1/43.0 \\
12 & 50\% & 78.7/42.8 \\
\cmidrule(lr){1-3}
10 & 25\% & 78.7/42.1 \\
10 & 75\% & 78.2/42.4 \\
\bottomrule
\end{tabular}

\end{wraptable}

\paragraph{spatial partition number $N$ and top-$K$ section ratio.}To assess the sensitivity of SME to its spatial configuration, we vary the spatial partition number $N$ and the top-$K$ selection ratio $K$ on the FLIR benchmark. All experiments use the full ADCR model. As reported in Tab.~\ref{tab:nk_sensitivity}, when fixing $K=50\%$, changing $N$ from 8 to 12 results in mAP$_{50}$/mAP$_{95}$ scores of 78.6/42.6, 79.1/43.0, and 78.7/42.8, respectively. The default setting $N=10$ achieves the best performance, suggesting that it provides a suitable balance between spatial granularity and contextual coverage. We further fix $N=10$ and vary $K$ from 25\% to 75\%. The performance
remains competitive, with scores of 78.7/42.1 and 78.2/42.4 for $K=25\%$ and $K=75\%$, respectively, compared with 79.1/43.0 for the default $K=50\%$. These results indicate that SME is not highly sensitive to moderate changes in its spatial partition or selection
ratio, while the balanced configuration $(N=10,K=50\%)$ provides the best overall detection accuracy and is therefore used in the remaining experiments.

\subsection{Visualization Results}

\begin{figure*}[htb!]
\begin{center}
\includegraphics[width=1\linewidth]{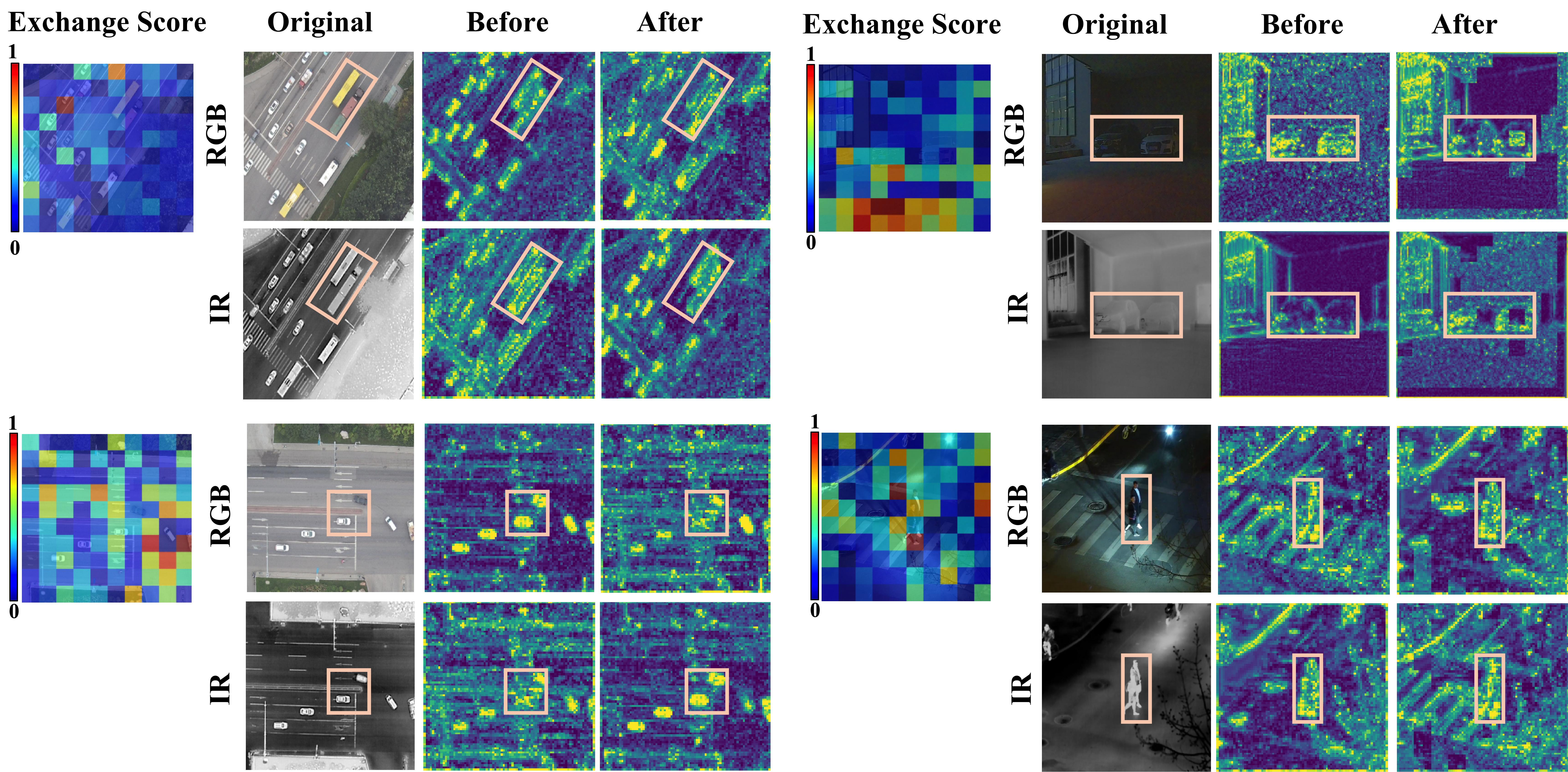}  
\end{center}
   \caption{\textbf{Visualization of SME mechanism.} The labeled boxes represent the locations of key targets in figures.
}
\label{fig:visual}
\end{figure*}

We visualize the outcomes of SME across four 
scenes from multiple datasets. As shown in \textbf{Fig.}~\ref{fig:visual}, For each scene, we present a block-level exchange score map, the original RGB and infrared(IR) image pairs, and the feature maps before and after SME is applied.

The exchange score map reflects the spatial distribution of complementarity between modalities:  higher-score regions where the two modalities exhibit greater semantic divergence and thus higher exchange priority. The exchange score map identifies blocks of maximum cross-modal semantic divergence, which are selected for bidirectional feature exchange regardless of whether they correspond to foreground objects or background regions. As shown in the highlighted areas, the exchange produces two 
distinct observable effects: in some regions, previously ambiguous activations become more structured after receiving features from the complementary modality, while the donor branch correspondingly absorbs the divergent context; in others, modality-specific noise 
patterns are redistributed between branches rather than suppressed. SME does not remediate a weak modality with a strong one, but instead exposes each backbone branch to 
cross-modal feature patterns it would not encounter 
under normal training, preventing over-specialization 
to single-modality statistics. The downstream LCC module consequently faces a harder channel arbitration problem during training, where modality provenance is deliberately obfuscated, making the network more robust to the modality asymmetry encountered at inference.

\begin{figure*}[htb!]
\begin{center}
\includegraphics[width=1\linewidth]{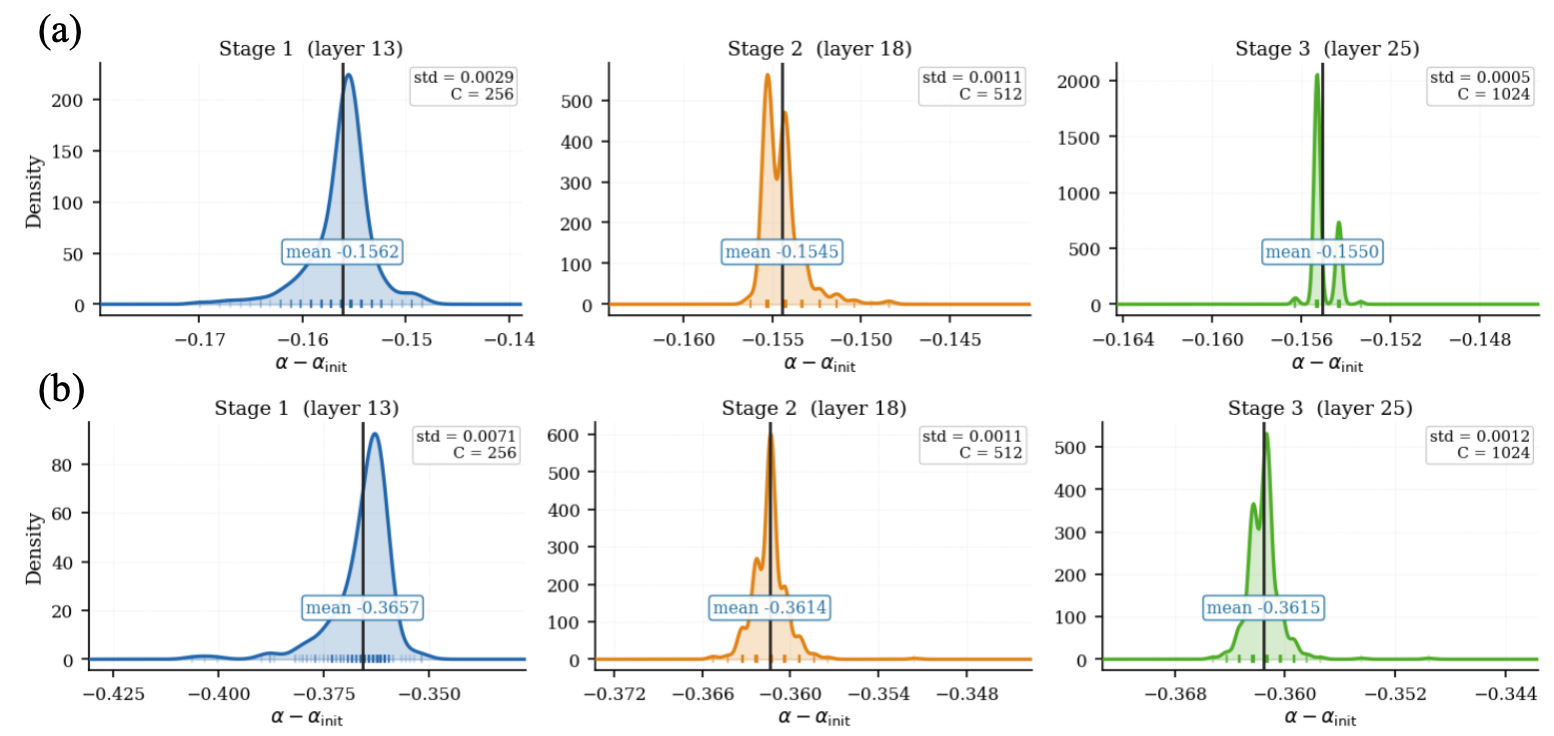}  
\end{center}
   \caption{\textbf{Per channel $\alpha$ deviation from initialization.} (a) is the result of VEDAI, (b) is the result of FLIR. $\alpha - \alpha_{init}$ is the offset of learnable parameter $\alpha$.
} 
\label{fig:alpha}
\end{figure*}

\textbf{Fig.}~\ref{fig:alpha} shows that per-channel $\alpha$ deviations from initialization become more concentrated in deeper stages, indicating more uniform surrogate mixing weights. FLIR exhibits a larger negative mean shift than VEDAI (\(-0.362\) vs. \(-0.156\)), corresponding to a stronger sigmoid contribution, alongside greater cross-channel variation at Stage 1 (std: \(0.0071\) vs. \(0.0029\)). These patterns indicate dataset- and stage-dependent adaptation of the surrogate mixture.

These learned mixtures determine how detection-loss gradients propagate through channel arbitration. For a given channel-score difference, the clipped-linear component concentrates its gradient within a bounded region around the decision boundary, whereas the sigmoid component also supplies gradients outside that region. A larger sigmoid weight thus redistributes the surrogate gradient toward score differences at which the clipped-linear component has saturated, allowing established channel preferences to remain adjustable. Through the channel-attention module, this gradient shaping influences updates to the preceding backbone and consequently the representations used for multimodal detection.

This mechanism also connects LCC to SME during training. SME exchanges spatial features between branches, changing the feature contrast that contributes to the arbitration gradient. The learned $\alpha$ controls how this task feedback propagates through the channel-score difference. In turn, updates to the shared feature extractor modify the attention gaps and semantic distances used by SME in subsequent training iterations, influencing which regions are selected for exchange. Together, these observations and the gradient mechanism illustrate how learnable $\alpha$ adapts channel-arbitration training to dataset-specific feature statistics while interacting with SME through the evolving feature representations.

\section{Conclusion}
This paper addresses the fundamental challenge of multimodal fusion under modality asymmetry, where direct feature mixing leads to gradient ambiguity when one modality is degraded by environmental conditions. In the context of RGB-infrared object detection, we identify that existing fusion paradigms face an inherent optimization conflict: a single fusion function cannot simultaneously act as a soft integrator under balanced modalities and a hard isolator under asymmetric ones. The proposed solution decouples this conflict into two complementary mechanisms operating at different stages. As opposed to previous work that performs feature mixing directly, SME circulates semantically divergent spatial features between backbone branches during training, preventing modality over-specialization without introducing additional supervision signals. LCC then performs channel 
arbitration  at inference via a differentiable competition mechanism, selecting the more informative modality per channel. Together, the two modules form a plug-and-play training-inference 
decoupled fusion paradigm.
Experiments on five benchmark datasets demonstrate continuous performance improvements. The most significant improvements occur on datasets with severe modal asymmetry and spatial misalignment. Future research may include addressing temporal modal asymmetry in video sequences and exploring interaction mechanisms for more modal inputs. We hope to further extend the method to more multimodal tasks.

\section{Acknowledgment}
Computational resources are supported by the National Academic Infrastructure for Super-computing in Sweden (NAISS) with Project No. NAISS 2026/4-674.

\bibliography{egbib}

\end{document}